\documentclass{article}

\usepackage{arxiv}

\usepackage[utf8]{inputenc} % allow utf-8 input
\usepackage[T1]{fontenc}    % use 8-bit T1 fonts
\usepackage{hyperref}       % hyperlinks
\usepackage{url}            % simple URL typesetting
\usepackage{booktabs}       % professional-quality tables
\usepackage{amsfonts}       % blackboard math symbols
\usepackage{nicefrac}       % compact symbols for 1/2, etc.
\usepackage{microtype}      % microtypography
\usepackage{xcolor}         % colors        
\usepackage{amsmath}
\usepackage{enumitem}
\usepackage{graphicx} 
\usepackage{amsthm}
\usepackage{textcomp}
\newtheorem{theorem}{Theorem}
\usepackage{amssymb}
\usepackage{footnote}
\usepackage{graphicx}

\usepackage[style=numeric, maxcitenames=1, mincitenames=1]{biblatex}
 
\title{Technical Report: Quantifying and Analyzing the Generalization Power of a DNN}

\author{
 Yuxuan He \\
  University of Electronic Science and Technology of China\\
  \texttt{youcannnnt@gmail.com} \\
  \And
 Junpeng Zhang \\
  Shanghai Jiao Tong University\\
  \texttt{zhangjp63@sjtu.edu.cn} \\
  \And
  Lei Cheng \\
  Shanghai Jiao Tong University\\
  \texttt{chenglei20020408@sjtu.edu.cn}
  \And
 Hongyuan Zhang \\
  Institute of Artificial Intelligence, China Telecom\\
  \texttt{hyzhang98@gmail.com} \\
  \And
 Quanshi Zhang \\
  Shanghai Jiao Tong University\\
  \texttt{zqs1022@sjtu.edu.cn} \\
}

\begin{document}

\maketitle

\begin{abstract}
This paper proposes a new perspective for analyzing the generalization power of  deep neural networks (DNNs), \emph{i.e.}, directly disentangling and analyzing the dynamics of generalizable and non-generalizable interaction encoded by a DNN through the training process. Specifically, this work builds upon the recent theoretical achievement in explainble AI \cite{ren2023we}, which proves that the detailed inference logic of DNNs can be can be strictly rewritten as a small number of AND-OR interaction patterns. Based on this, we propose an efficient method to quantify the generalization power of each interaction, and we discover a distinct three-phase dynamics of the generalization power of interactions during training. In particular, the early phase of training typically removes noisy and non-generalizable interactions and learns simple and generalizable ones. The second and the third phases tend to capture increasingly complex interactions that are harder to generalize. Experimental results verify that the learning of non-generalizable interactions is the the direct cause for the gap between the training and testing losses.
\end{abstract}

\section{Introduction}
\textcolor{blue}{This paper is just a technical report for the quantification of the generalization power of a DNN. The release of arXiv technical report is just for the citation and does not hurt the novelty of the research paper.}

Despite the rapid development of deep learning, the generalization power of a deep neural network (DNN) has not been sophisticatedly explained. Most widely used methods of enhancing a DNN’s generalization power are empirical essentially, \emph{e.g.}, chain of thoughts \cite{COT} , data cleaning \cite{,brown2020language,noiserobust} and augmentation \cite{cubuk2019autoaugment,dmgan}, and LLM alignment \cite{rlhf,dpo}. Meanwhile, there remains a salient gap between application requirements and theoretical analysis of the generalization power \cite{Foret_Kleiner_Mobahi_Neyshabur_2020,landscape,xiao2020disentangling}.

Therefore, in this paper, we consider two new issues regarding the generalization power of DNNs: \textit{(1) Can the intricate inference logic of a DNN on an input sample be reformulated in mathematics as a set of symbolic concepts? (2) Can we leverage such concepts to explain the generalization power of DNN's?} \textbf{If so, we can track the change of DNN's symbolic concepts during the training process to explain the dynamics of the DNN's generalization power in mathematics.}

\textbf{Background.} While interpreting DNNs' complex inference logic through symbolic concepts remains a key goal of explainable AI, symbolic explanations of neural networks have long been considered counterintuitive and fundamentally unattainable in mathematics \cite{rudin2019stop}. \textbf{Fortunately, Ren et al. \cite{ren2023we} have proven a series of theorems, which guarantee that  using interaction concepts can accurately explain all the detailed inference logic of DNNs.} Specifically, given an input sample, each interaction represents an AND relationship or an OR relationship between input variables encoded by the DNN, and it makes a certain numerical effect on the output. For example, an interaction \( S = \{\text{Good morning}\} \) encoded by an LLM usually represents a phrase automatically learned by the LLM, and pushes the output towards polite or greeting-related semantics. A logical model based on a set of AND-OR interaction logics can accurately predict the DNN’s outputs on numerous randomly augmented input samples. This has been widely regarded \cite{liu2023towardsdiffc,ren2021can,ren2023bayesian,zhou2024explaining} as the theoretical foundation for the emerging direction of interaction-based explanation.

\textbf{Our research.} In this paper, we find that the above theory enables us to concretize the generalization power of a DNN. Since the DNN's output score can be reformulated as the sum of all interaction effects, if most interactions learned by the DNN can also frequently appear in (be transferred to) unseen testing samples, then the DNN would exhibit a high testing accuracy. In other words, \textbf{the generalization power of interactions determines the generalization of the entire DNN.} For example, in bird species classification, the generalizable interaction between image regions of \( \{ \text{red features}, \text{long beak}\} \) can be generalized (transferred) to unseen testing data and push the DNN output towards the \textit{Flamingo} category.

To this end, previous studies \cite{zhou2024explaining} could only analyze the generalization power of a DNN by quantifying the similarity of the interaction distributions between training data and testing data. Up to now, there still lacks an efficient method to quantify the generalization power of each specific interaction. Therefore, we propose a efficient method to quantify each specific interaction's generalization power. We train a baseline DNN and measure the proportion of interactions in a DNN that can be transferred to a baseline DNN trained on the testing set.

\begin{figure}[t]  % h 表示尽量将图片放置在此处
    \centering
    \includegraphics[width=\textwidth]{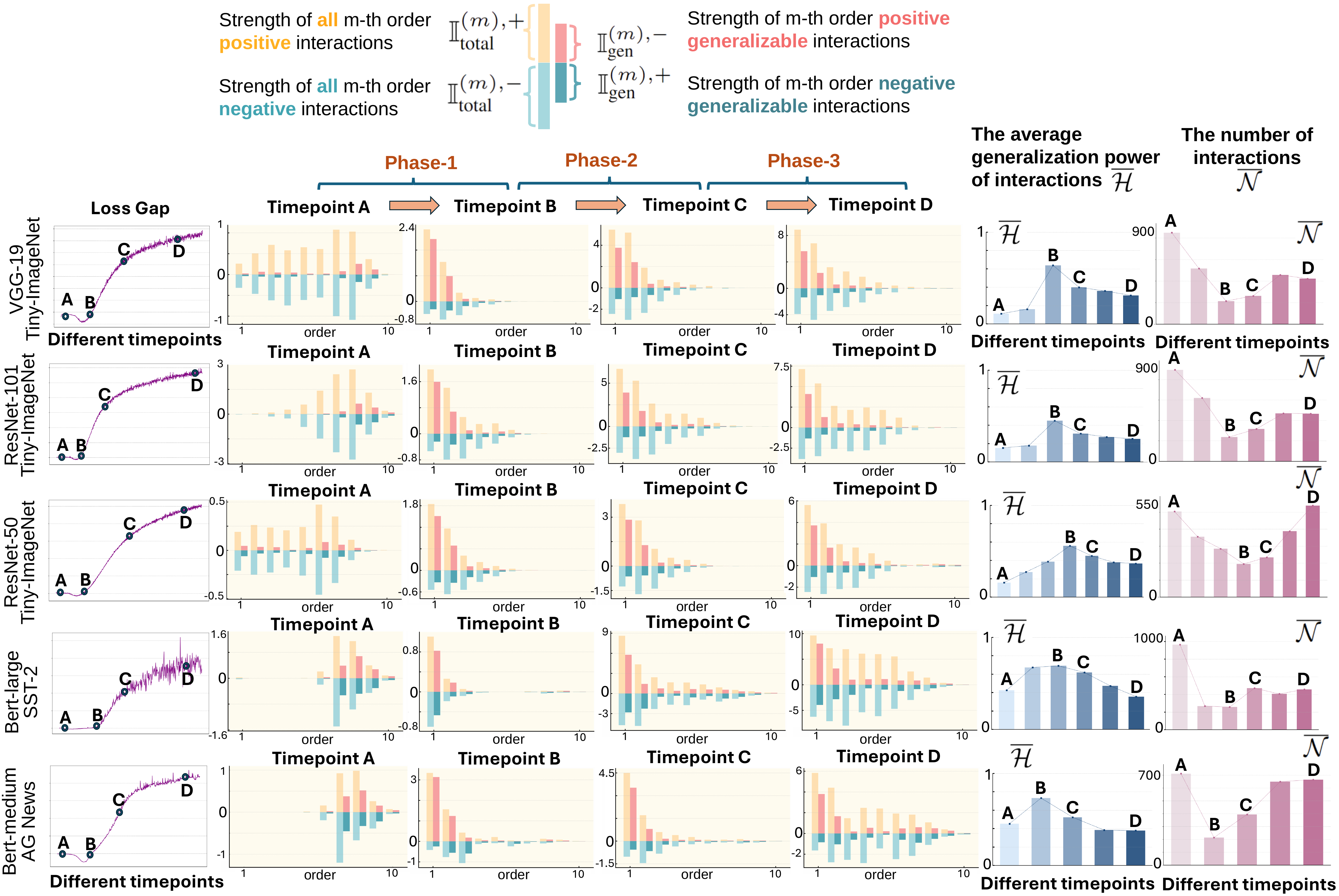}  % 插入图片，width 控制图片宽度
    \caption{We visualize the change of interactions through the entire training process and track the change of the average generalization power of interactions $\overline{\mathcal{H}}$ and the number of interactions $\overline{\mathcal{N}}$. The dynamics of quantity and the average generalization power of interactions explains the change of the loss gap. }  % 图片标题
    \label{fig:Three-phase-detail}  % 图片标签，用于引用
\end{figure}

This new evaluation method allows us to first uncover the distinctive three-phase dynamics of the generalization power of interactions in the DNN through the entire training process (see Figure \ref{fig:Three-phase-detail}):
\begin{itemize}[label={\large$\bullet$},left=0cm] 
  \item \textbf{Phase 1:} Given a fully initialized DNN, in the early epochs of training, a large number of non-generalizable interactions are removed from the DNN, while a few generalizable interactions are gradually learned. These generalizable interactions are often simple and involve a small number of input variables, leading to the enhancement of the DNN's generalization power.
  \item \textbf{Phase 2:} The DNN continues to learn more interactions, but the newly learned interactions exhibit increasing complexity (\emph{i.e.}, learning interactions between more input variables). These more complex interactions often have poorer generalization power. Thus, although the testing loss of the DNN continues to decreases, it tends to saturate due to the decreasing generalization power of the newly learned interactions.
  \item \textbf{Phase 3:} The DNN continues to learn additional interactions, but these newly learned interactions are often complex and difficult to apply to unseen data. Very few of these late-stage learned interactions represent simple, broadly generalizable patterns. Therefore, we can consider this phase revealing the overfitting of the DNN. During this phase, the gap between training and testing loss continues to widen. And the testing loss may even increase.
\end{itemize}

\section{Methodology}
\subsection{Preliminaries: AND-OR Interaction}\label{subsec:pre}

Consider an input sample $\mathbf{x} = [x_1, x_2, \dots, x_n]^T$ comprising $n$ input variables, where $N = \{1, 2, \ldots, n\}$ denotes the index set. Let $v(\mathbf{x}) \in \mathbb{R}$ denote the scalar output of the DNN. While $v(\mathbf{x})$ can be defined in various ways\footnote{For instance, $v(\mathbf{x})$ can be set to the feature dimension of the ground-truth category prior to the softmax operation.}, we adopt the standard formulation widely used by \cite{ren2023we, zhou2024explaining, liu2023towardsdiffc} to let $v(\mathbf{x})$ represent the classification confidence score for the ground-truth label $\mathbf{y^*}$ as
\begin{equation}
v(\mathbf{x}) = \log \left( \frac{p(\mathbf{y^*} \mid \mathbf{x})}{1 - p(\mathbf{y^*} \mid \mathbf{x})} \right)\text{.}
\label{eq:out_score}
\end{equation}
\noindent Theorem 1 establishes that for a given DNN $v$, there exists a corresponding surrogate logical model $d(\cdot)$ based on AND-OR interactions that precisely captures and explains the complete inference logic of the original DNN $v$:
\begin{equation}
\!\!\!d(\mathbf{x_{\text{mask}}}) =\!\!\!\!\! \sum_{T \in \Omega_{\text{and}}}  \underbrace{I_T^{\text{and}} \cdot \delta_{\text{and}}\left(\ \substack{\mathbf{x}_{\text{mask}}\ \text{triggers AND relation}\\ \text{between input variables in}\ T} \right)}_\text{an AND interaction between input variables in T} +\!\!\! \sum_{T \in \Omega_{\text{or}}} \underbrace{I_T^{\text{or}} \cdot \delta_{\text{or}}\left(\ \substack{\mathbf{x}_{\text{mask}}\ \text{triggers OR relation}\\ \text{between input variables in}\ T} \right)}_\text{an OR interaction between input variables in T} \!+\ b,
\label{eq:universal_matching}
\end{equation}
\noindent where $\mathbf{x_{\text{mask}}}$ denotes the input $\mathbf{x}$ in which some input variables are masked\footref{fn3} and, $b$ is a scalar bias.

\textbf{Each AND interaction} is represented by a binary function $\delta_{\text{and}}\left(\ \substack{\mathbf{x}_{\text{mask}}\ \text{triggers AND relation}\\ \text{between input variables in}\ T} \right) \in \{0, 1\}$. It returns $1$ if all variables in $T$ are not masked in $\mathbf{x}_{\text{mask}}$; otherwise it returns $0$. $I_T^{\text{and}}$ is a scalar weight.

\textbf{Each OR interaction} is represented by a binary function $\delta_{\text{or}}\left(\ \substack{\mathbf{x}_{\text{mask}}\ \text{triggers OR relation}\\ \text{between input variables in}\ T} \right) \in \{0, 1\}$. It returns $1$ if any variables in $T$ are not masked in $\mathbf{x}_{\text{mask}}$; otherwise it returns $0$. $I_T^{\text{or}}$ is a scalar weight.

\textbf{The order of an interaction $S$} represents the complexity of the interaction. The order is defined as the number of input variables involved in the interaction $\text{Order}_S=\vert S\vert$.

\textbf{The universal matching property} \textit{in Theorem \ref{thm:universal_matching_property} demonstrates that the constructed logical model is able to accurately predict all outputs of the DNN across an exponential number of masked input states. Thus, this assures that the DNN can be regarded as equivalently utilizing AND-OR interactions as its primitive inference patterns for classification.}
\begin{theorem}[\textbf{universal matching property}, proven by \cite{chen2024extractgeninter}]
Given DNN \( v \) and an input sample \( \mathbf{x} \), let us set the scalar weights as \(\forall T \subseteq N\text{,}\) \(
I^{\text{and}}_T = \sum_{L \subseteq T} (-1)^{|T| - |L|} o^{\text{and}}_L \text{,}  \ I^{\text{or}}_T = - \sum_{L \subseteq T} (-1)^{|T| - |L|} o^{\text{or}}_{N \setminus L} \  \textbf{subject to} \  o^{\text{and}}_L + o^{\text{or}}_L = v(\mathbf{x}_L)\), and let $b=v(\emptyset)$\footnote{$v(\emptyset)$ represents the network output when all input variables in \(\mathbf{x}\) are masked.}. The logical model can then accurately predict the network outputs as follows, regardless of how \( \mathbf{x} \) is randomly masked:
\begin{equation}
\forall S \subseteq N, \quad d(\mathbf{x}_S) = v(\mathbf{x}_S),
\end{equation}
\noindent{where \( \mathbf{x}_S \) represents a masked input \(\mathbf{x}\), in which input variables in \(N \setminus S\) are masked\footnote{\( \mathbf{x}_S \) is generated by substituting all input variables in \( N \setminus S \) with a baseline value. 
We set the baseline value of an input variable as the average value of this variable across different input samples. For NLP models, we use a specific embedding in \cite{cheng2024layerwise,ren2023we} to mask input tokens.}
 \label{fn3}}.
\label{thm:universal_matching_property}
\end{theorem}
\textbf{Conciseness of the logical model.} Ren et al. \cite{ren2023we} have further proven that the logical model can be very concise, \emph{i.e.,} for most well-trained DNNs, the logical model can contain only a small number ($\mathcal{O}\left( \frac{n^{p}}{\tau} \right) \ll 2^{n+1}$ of nodes of AND-OR interactions with salient effects, while all other interactions have almost zero effects. Empirically, $p \in [0.9, 1.2]$. This conciseness enables a simpler logical model with a few salient AND-OR interactions in $\Omega_\text{and}^\text{Salient} = \{ S \mid \lvert I_S^{\text{and}} \rvert > \tau \}$ and $\Omega_\text{or}^\text{Salient} = \{ S \mid \lvert I_S^{\text{or}}\rvert > \tau \}$ to provide a concise and faithful explanation of the inference patterns in the DNN as follows:
\begin{equation}
\forall S \subseteq N, \quad v(\mathbf{x}_S) \approx d'(\mathbf{x}_S),
\end{equation}
\begin{equation}
d'(\mathbf{x}_S) =\!\!\!\!\!\sum_{T \in \Omega_\text{and}^\text{Salient}}\!\!\! I_T^{\text{and}} \cdot \delta_{\text{and}}\left(\ \substack{\mathbf{x}_{\text{mask}}\ \text{triggers AND relation}\\ \text{between input variables in}\ T} \right)+\!\!\!\!\!\sum_{T \in \Omega_\text{or}^\text{Salient}} \!\!\!I_T^{\text{or}} \cdot \delta_{\text{or}}\left(\ \substack{\mathbf{x}_{\text{mask}}\ \text{triggers OR relation}\\ \text{between input variables in}\ T} \right) + b\text{.}
\label{eq:universal_sal}
\end{equation}
\textbf{Computation of interactions.} 
Building upon the framework established in [3, 17], we reformulate $o^{\text{and}}_S=0.5 \cdot v(\mathbf{x}_S)+\gamma_S$ and $o^{\text{or}}_S=0.5 \cdot v(\mathbf{x}_S)-\gamma_S$ in Theorem \ref{thm:universal_matching_property}, which satisfy the required relation $v(\mathbf{x}_S) = o^{\text{and}}_S + o^{\text{or}}_S$. The parameters $\{\gamma_S\}$ are then optimized by minimizing a LASSO-regularized loss $\min_{\{\gamma_T\}} \left\| \mathbf{I}_{\text{and}} \right\|_1 + \left\| \mathbf{I}_{\text{or}} \right\|_1$ to obtain the sparsest interaction patterns, where $\mathbf{I}_{\text{and}} = \left[ I^{\text{and}}_{S_1}, \dots, I^{\text{and}}_{S_{2^n}} \right]^T, \ \mathbf{I}_{\text{or}} = \left[ I^{\text{or}}_{S_1}, \dots, I^{\text{or}}_{S_{2^n}} \right]^T \in \mathbb{R}^{2^n}$.

\subsection{Quantifying generalization power of interactions}\label{sec:quantify}
The above universal matching property and the sparsity property have become foundational pillars for the emerging direction of interaction-based explanation \cite{li2023does,ren2023bayesian,zhou2024explaining}. Crucially, interaction theory introduces a paradigm shift in understanding the generalization behavior of DNNs. This is grounded in the observation that a DNN's predictive output can be mathematically decomposed into a linear combination of AND-OR interaction effects. Consequently, a DNN's generalization power can be considered to be determined by the generalization power of the interactions in the DNN.

\textbf{Previous definition of the generalization power of interactions.} Let us first revisit the definition of generalization power of an interaction in \cite{zhou2024explaining}. Given a salient AND interaction $S$ extracted from an input $\mathbf{x}$ subject to $|I^{\text{and}}_S| > \tau$, if this interaction $S$ frequently occurs in the testing set and consistently making an effect on the classification of a category, then it is deemed to be generalizable; otherwise, it fails to generalize to testing samples. This criterion extends analogously to OR interactions. For instance, in a bird species classification task, suppose an AND interaction $S=\{\text{red features},\text{long beak}\}$ consistently appears in both training and testing datasets and consistently increases the confidence in predicting the "Flamingo" class. In this case, the interaction $S$ demonstrates strong generalization power to unseen testing samples.

\textbf{Efficient quantification of the generalization power of interactions.} However, the above definition does not provide an efficient method for quantifying the generalization power of interactions. For example, in natural language processing tasks, this would necessitate an exhaustive search for specific interactions across numerous testing samples, leading to prohibitive computational costs.

Therefore, we introduce an approximate and efficient approach. Instead of performing an exhaustive search, we evaluate the transferability of an interaction $S$ encoded by the DNN to a baseline DNN trained on the testing samples. The successful transfer to baseline DNN of the interaction $S$ is regarded as compelling evidence that the interaction $S$ captures a fundamental pattern within the testing samples, thus strongly suggesting its generalization power.

To elucidate this approach, let us consider an extracted salient AND interaction $S$, such that $|I^{\text{and}}_S| > \tau$. A distinct \textit{baseline} DNN, denoted by $v^{\text{base}}$, is subsequently trained on the testing samples. To this end, the interaction $S$ is deemed to represent a generalizable pattern within the testing set if the baseline DNN $v^{\text{base}}$ concurrently identifies the AND interaction $S$ to have salient effect (\emph{i.e.}, $|I^{\text{and}}_{S,v^{\text{base}}}| > \tau$) and the AND interaction $S$ exerts a consistent directional influence on the classification of $\mathbf{x}$ (\emph{i.e.}, $I^{\text{and}}_{S,v^{\text{base}}} \cdot I^{\text{and}}_S > 0$). The same principle applies to OR interactions. This approach is predicated on the assumption that the baseline DNN captures the inherent interactions of the testing set (see Appendix C for ablation study). Accordingly, the generalization power of an AND/OR interaction is assessed by the following binary metrics:
\begin{equation}
\begin{split}
    \mathcal{G}^{\text{and}}_{S,v^{\text{base}}} &= \mathbf{1}(\vert I^{\text{and}}_{S,v^{\text{base}}} \vert > \tau) \cdot \mathbf{1}(I^{\text{and}}_S \cdot I^{\text{and}}_{S,v^{\text{base}}} > 0) \in \{0, 1\}, \\
    \mathcal{G}^{\text{or}}_{S,v^{\text{base}}} &= \mathbf{1}(\vert I^{\text{or}}_{S,v^{\text{base}}} \vert > \tau) \cdot \mathbf{1}(I^{\text{or}}_S \cdot I^{\text{or}}_{S,v^{\text{base}}} > 0) \in \{0, 1\},
\label{eq:gen_inter}
\end{split}
\end{equation}
\noindent where $\mathbf{1}(\cdot)$ is a binary indicator function that outputs 1 only when the the condition is satisfied.

\printbibliography
\end{document}